\documentclass[10pt,twocolumn,letterpaper]{article}

\usepackage{cvpr}              

\usepackage[dvipsnames]{xcolor}

\definecolor{cvprblue}{rgb}{0.21,0.49,0.74}
\usepackage[pagebackref,breaklinks,colorlinks,citecolor=cvprblue]{hyperref}
\usepackage{pifont}
\usepackage{tabularx}
\usepackage{multirow}
\newcommand{\model}{\textsc{FastMesh}\xspace}
\newcommand{\cmark}{\ding{51}}
\newcommand{\xmark}{\ding{55}}

\newcommand{\revise}[1]{{#1}}

\definecolor{lightgreen}{RGB}{204, 255, 204}
\definecolor{lightblue}{RGB}{204, 229, 255}

\title{\model: Efficient Artistic Mesh Generation via Component Decoupling}

\author{Jeonghwan Kim\quad Yushi Lan\quad Armando Fortes\quad Yongwei Chen\quad Xingang Pan\\
S-Lab, Nanyang Technological University \\
\href{https://jhkim0759.github.io/projects/FastMesh/}{https://jhkim0759.github.io/projects/FastMesh}
}

\begin{document}
\twocolumn[{
	\renewcommand\twocolumn[1][]{#1}
	\maketitle
	\vspace{-9mm}
	\begin{center}
		\includegraphics[width=\textwidth]{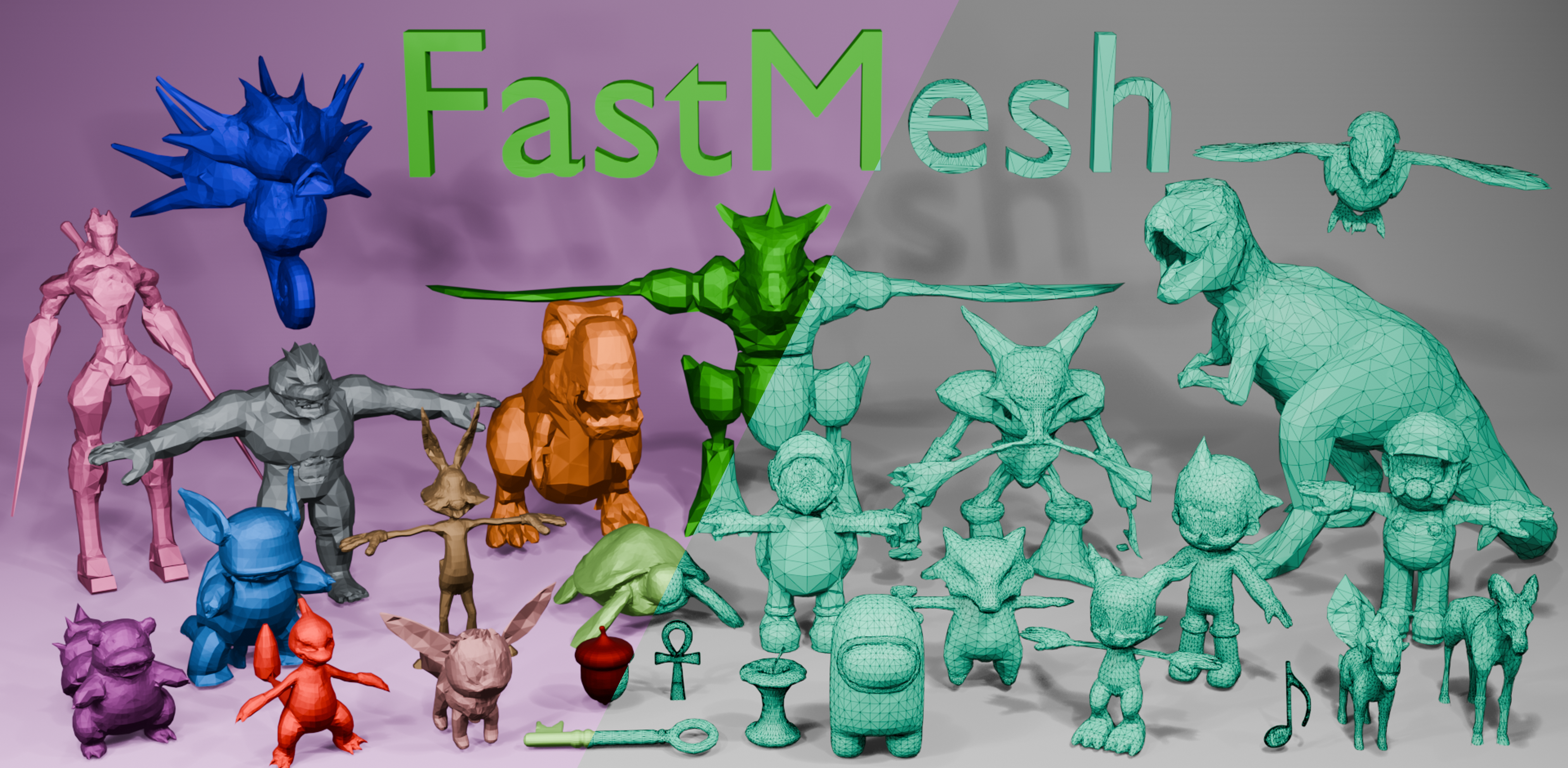}
	\end{center}
        \vspace{-5mm}
	\captionof{figure}{Example of meshes generated by \model. Our approach efficiently produces 3D objects by substantially reducing the number of tokens required for generation. Note that all meshes are directly generated from point clouds.}
        \vspace{5mm}
	\label{fig:teaser}
    \vspace{-1mm}
}]

\begin{abstract}
Recent mesh generation approaches typically tokenize triangle meshes into sequences of tokens and train autoregressive models to generate these tokens sequentially. 
Despite substantial progress, such token sequences inevitably reuse vertices multiple times to fully represent manifold meshes, as each vertex is shared by multiple faces. This redundancy leads to excessively long token sequences and inefficient generation processes. In this paper, we propose an efficient framework that generates artistic meshes by treating vertices and faces separately, significantly reducing redundancy. We employ an autoregressive model solely for vertex generation, decreasing the token count to approximately 23\% of that required by the most compact existing tokenizer. Next, we leverage a bidirectional transformer to complete the mesh in a single step by capturing inter-vertex relationships and constructing the adjacency matrix that defines the mesh faces. To further improve the generation quality, we introduce a fidelity enhancer to refine vertex positioning into more natural arrangements and propose a post-processing framework to remove undesirable edge connections. Experimental results show that our method achieves more than 8$\times$ faster speed on mesh generation compared to state-of-the-art approaches, while producing higher mesh quality.
\end{abstract}
\vspace{-5mm}
\section{Introduction}
Industries such as gaming, visual effects, and virtual reality rely heavily on 3D meshes as the core representation, due to their compactness and compatibility with mature rendering pipelines. As the demand for 3D assets continues to grow, creating high-quality meshes through traditional manual modeling remains time-consuming and labor-intensive. This has driven a wave of research into data-driven generative models that aim to automate mesh creation, either from scratch or conditioned on point clouds~\cite{Siddiqui2024meshgpt,chen2024meshxl,chen2024meshanything,gao2024meshart,hao2024meshtron,wang2024llamamesh,weng2024pivotmesh}. These approaches typically represent a mesh as a sequence of tokens and train an LLM–style architecture to generate the tokens autoregressively.

Despite promising advances in mesh generation, the next-token prediction paradigm faces a key limitation: it is inefficient in both time and memory due to the long token sequences. In a mesh, multiple faces often share common vertices. As a result, tokenization requires repeatedly recording the same vertices across different tokens, leading to significant redundancy. Consequently, these methods typically take 30 seconds to one minute to generate a mesh with 500 vertices on an A6000 GPU, and they struggle to scale to complex meshes with a large number of vertices. While some works have proposed more efficient tokenization strategies to improve compression ratios~\cite{chen2024meshanythingv2,weng2024scaling,lionar2025treemeshgpt,wang2025nautilus}, repeated references to the same vertex tokens remain unavoidable when encoding geometric information into a single sequence.

In this paper, we introduce \model, an efficient framework for generating high-quality 3D meshes within seconds.
The key idea is to decouple the generation of mesh components—vertices and faces—and process them sequentially to avoid duplication issues commonly encountered in mesh tokenization.
For vertex generation, we employ an autoregressive model, as it inherently accommodates varying vertex counts.
Once vertices are generated, we recognize that their connectivity primarily depends on local information, allowing for parallel processing.
To facilitate this, we leverage a bidirectional transformer~\cite{vaswani2017attention} to model the relationships between vertices, from which edge connections can be directly derived in one step.
These edges form an adjacency matrix that is then used to extract faces by identifying closed triangles.
Thanks to this design, \model reduces the token count to approximately 23\% of that required by the previous most compact tokenizer~\cite{weng2024scaling}, significantly mitigating issues associated with long token sequences, such as high inference latency and quality degradation.

Beyond this, we introduce a fidelity enhancer module and a prediction filtering process to further improve mesh quality. 
The fidelity enhancer maps discretized vertex positions, which are constrained by indexing~\cite{weng2024scaling}, back to continuous coordinates. 
This results in smoother surfaces and more natural vertex distributions. 
In addition, the prediction filtering, a post-processing technique for face generation, refines the adjacency matrix by progressively masking irrelevant connections. 
This reduces redundant or spurious faces while preserving the intended geometric structure, yielding meshes that are cleaner, more compact, and better aligned with downstream requirements.

Extensive experiments on the Toys4K dataset~\cite{toys4K} demonstrate that \model can generate higher-quality 3D meshes with significantly less time than previous methods. 
When generating meshes with 1,000 and 4,000 vertices, our method only takes about 7 seconds and 30 seconds, respectively, achieving an $8\times$ speedup over BPT~\cite{weng2024scaling}.
In particular, since our mesh is generated in continuous space with a higher vertex count, it achieves not only greater geometric accuracy but also higher visual quality.
Furthermore, we show that our method can be effectively integrated with other 3D generation pipelines that produce non-mesh 3D assets, enabling the creation of artistic meshes aligned with diverse input modalities such as images or text. 
As illustrated in Fig.~\ref{fig:teaser}, \model accurately represents a diverse range of complex structures conditioned on shape inputs. 
Our contributions can be summarized as follows:
\begin{itemize}
    \item We propose an efficient framework for high-quality mesh generation, which treats vertices and faces separately, each with a dedicated model suitable for the task. In particular, once the vertices are ready, all the mesh edges are generated in parallel via a bidirectional transformer. Our design reduces the token count to about 23\% of that required by the previous most compact tokenizer~\cite{weng2024scaling}. 
    \item We integrate a fidelity enhancer to improve the precision of vertex representations by restoring information lost during quantization.
    Additionally, we propose prediction filtering to reduce connectivity errors while preserving geometric structure and mesh quality.
    \item \model shows a clear advantage through the Toys4K dataset by generating more detailed and accurate meshes at a significantly faster speed compared to prior methods.
\end{itemize}

\begin{figure*}
    \centering
    \centerline{\includegraphics[width=1.0\textwidth]{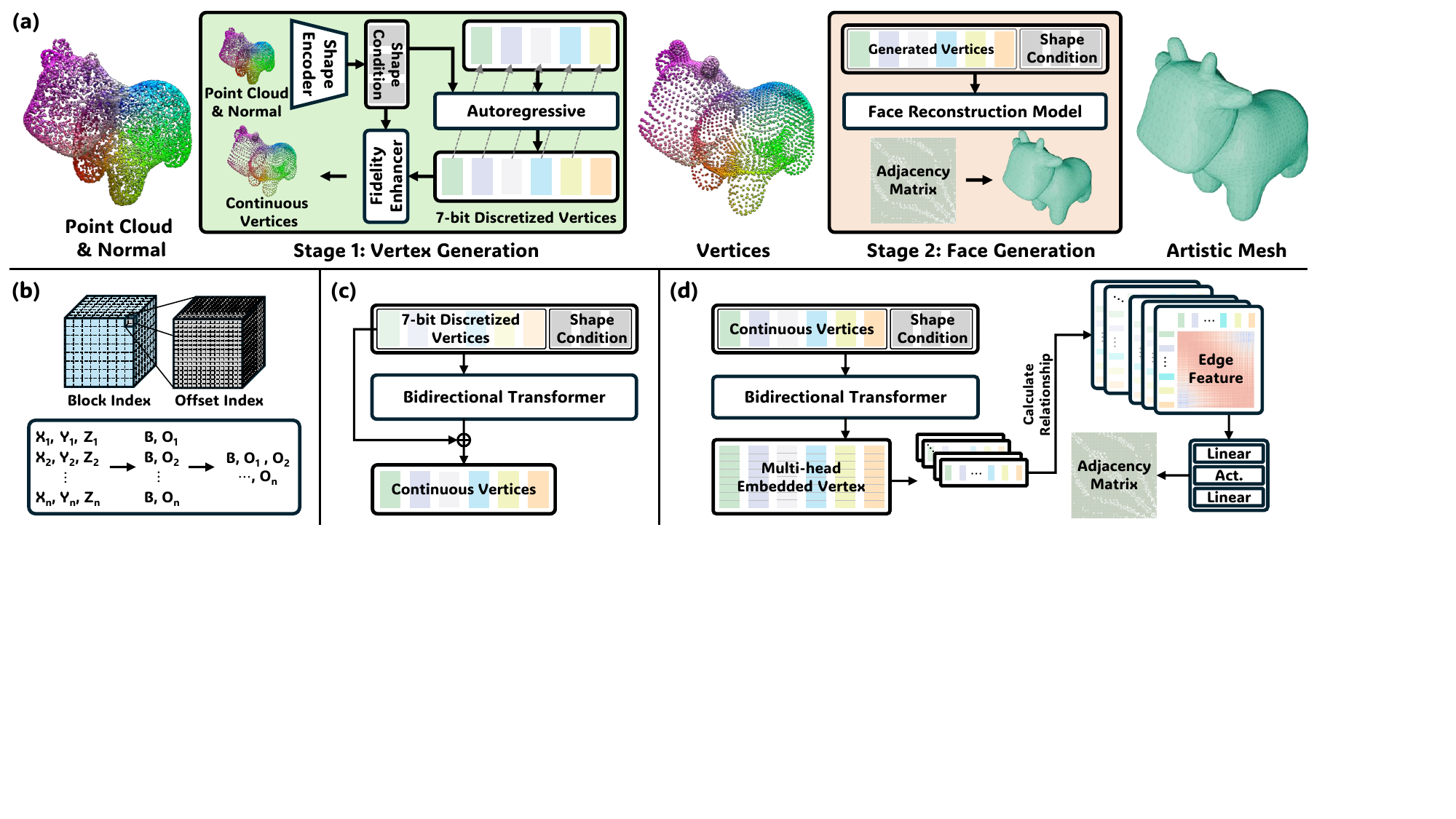}}
    \caption{\label{fig:pipeline}
    \textbf{(a)} Overall architecture of \model. Note that our pipeline consists of two stages, where we first generate the vertices from the shape condition and then construct the faces to complete the mesh. \textbf{(b)} Visualization of the block-wise indexing scheme introduced by BPT~\cite{weng2024scaling}, which we adopt for vertex tokenization.
    \textbf{(c)} Structure of the fidelity enhancer in the first stage. The 7-bit discretized vertices and shape condition are fed into the network to estimate the offset that can make the coordinate a continuous value. \textbf{(d)} Details of face reconstruction. The generated vertices are embedded to capture inter-vertex relationships in a multi-head manner. Each head computes a matrix, where the output represents one feature dimension used in edge prediction.}
    \vspace{-3mm}
\end{figure*}

\section{Related Work}
\subsection{3D Mesh Generation}
Early studies~\cite{mescheder2019occupancy,xie2019pix2vox,nichol2022point,vahdat2022lion,gao2022get3d,jun2023shap} on mesh generation, constrained by the scarcity of large-scale 3D datasets, trained on limited object categories, and often failed to generalize to out-of-distribution classes. DreamFusion~\cite{poole2023dreamfusion} pioneered the use of score distillation loss to train 3D models guided by diffusion priors~\cite{saharia2022photorealistic,rombach2022high}, inspiring a series of follow-up works~\cite{Lin2023magic3d,wang2023prolificdreamer,chen2023fantasia3d,tang2023make,chen2024comboverse} aimed at mitigating the lack of 3D data.
With the release of large-scale 3D datasets~\cite{objaverse,objaverseXL}, feed-forward methods~\cite{hong2023lrm,li2023instant3d,liu2023one,wang2024crm,xu2024instantmesh} demonstrated high-speed 3D generation, while subsequent diffusion-based approaches~\cite{liu2024one,lan2024ln3diff,lan2024ga,xiang2024structured,chen2025primx} produced more detailed and higher-quality outputs. Most of these methods utilize intermediate representations (\textit{e.g.}, point clouds, implicit functions, and triplane) rather than meshes to facilitate the training of 3D geometry. These are typically converted into meshes through post-processing techniques~\cite{bernardini2002ball,MarchingCube,shen2023flexicubes}, enabling use in downstream tasks.
However, such pipelines often yield dense meshes with overly smooth surfaces or cause distortions through interpolation, failing to faithfully represent the intended 3D structure.
Motivated by these limitations, a line of research has emerged on generating complete meshes end-to-end with neural networks. Notably, PolyGen~\cite{nash2020polygen}, MeshGPT~\cite{Siddiqui2024meshgpt}, and PolyDiff~\cite{alliegro2023polydiff} were among the first to leverage generative models for creating mesh triangles, producing results that closely resemble those crafted by human artists.

\subsection{Shape-conditioned Artistic Mesh Generation}
Since the initial works on artistic mesh generation were published, MeshXL~\cite{chen2024meshxl} and MeshAnything~\cite{chen2024meshanything} proposed the autoregressive model to create meshes by using shape information in the form of point clouds as input. It is worth noting that these designs enhance the scalability of artistic mesh generation, since all 3D representations can be easily converted into point clouds.
However, since each triangle requires three vertex coordinates—nine tokens in total—these approaches struggle as the sequence length grows with the number of faces. This makes it difficult for the network to handle complex meshes. Meshtron~\cite{hao2024meshtron} designed a hierarchical structure to efficiently process large amounts of tokens. 
More recent efforts~\cite{chen2024meshanythingv2,tang2025edgerunner,weng2024scaling,lionar2025treemeshgpt,wang2025nautilus} have proposed compact tokenization mechanisms that encode mesh geometry into shorter sequences while preserving topological consistency. Among them, BPT~\cite{weng2024scaling} introduces two promising schemes—block-wise indexing and patchified aggregation—achieving a compression rate of about 75\%. This substantial reduction in computational burden enables the network to process more complex and large-scale meshes with up to 8,000 faces. \revise{However, token compression alone does not fully resolve scalability—the sequence length remains lower-bounded by the number of faces, and discretization artifacts introduced by block-wise indexing persist.
On the other hand, SpaceMesh~\cite{spacemesh2024} adopts a diffusion model for vertex generation, followed by embeddings to capture inter-vertex relationships, and constructs faces by traversing edges in a half-edge representation. Although this approach can generate meshes within a few seconds, it suffers from low geometric accuracy and is restricted to low-poly outputs, primarily due to instability in predicting both vertices and edges. In contrast, \model is capable of generating high-complexity meshes with detailed structures while maintaining faster inference. This design not only overcomes the limitations of existing methods but also underscores the value of revisiting decoupled design principles.} 

\section{Proposed Method}
The proposed method addresses the inherent redundancy in vertex usage when tokenizing entire meshes as a single sequence, thereby facilitating faster inference and more effective handling of complex mesh structures. Similar to prior works~\cite{chen2024meshanythingv2, weng2024scaling, tang2025edgerunner}, we extract shape condition features from input point clouds and normals using a pretrained encoder~\cite{zhao2023michelangelo}. Subsequently, the network generates vertices and faces to complete the mesh. Accordingly, the overall architecture consists of two main stages, as shown in Fig.~\ref{fig:pipeline}(a).

\begin{figure}
    \centering
    \centerline{\includegraphics[width=1.0\linewidth]{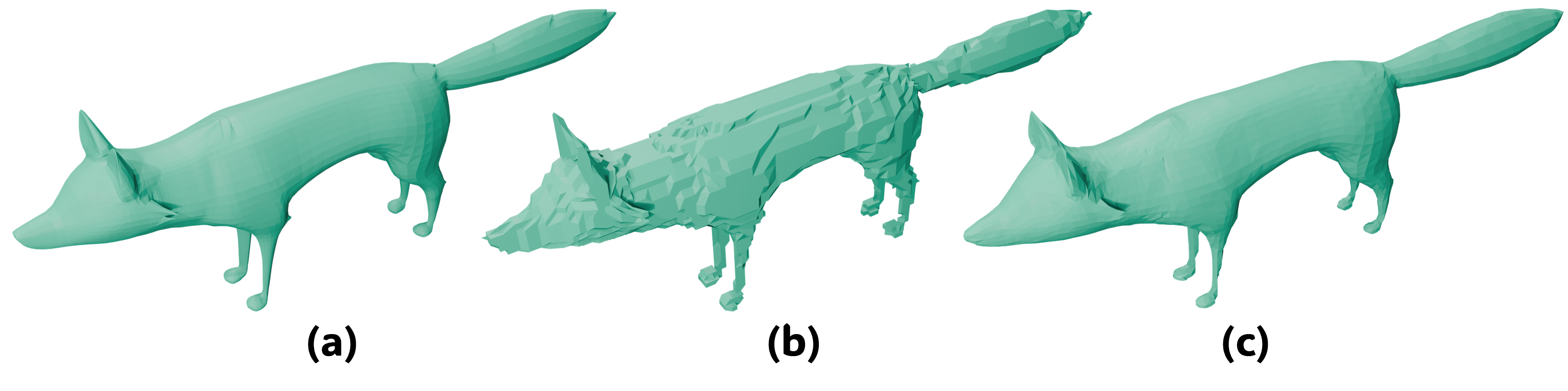}}
    \caption{\label{fig:discretize}
    \textbf{(a)} Example of high-resolution mesh containing 7,694 vertices. \textbf{(b)} Mesh obtained by discretizing (a) in 7-bit coordinate space, resulting in 3,636 vertices. \textbf{(c)} Mesh reconstructed using the same 3,636 vertices as in (b), with continuous coordinates refined by our fidelity enhancer.}
    \vspace{-3mm}
\end{figure}

\subsection{Stage \textrm{I}: Vertex Generation}
Since vertex generation significantly influences the overall mesh quality, we designed our approach to ensure high-quality vertex prediction. We first generate vertex sequences using an autoregressive model and then refine the vertex positions with the fidelity enhancer to achieve more precise and naturally arranged vertex distributions.
To support efficient generation in the autoregressive model, we employ the block-wise indexing method to compress the vertex representation sequence effectively.
This technique, originally introduced in  BPT~\cite{weng2024scaling}, is illustrated in Fig.~\ref{fig:pipeline}(b). Specifically, block-wise indexing maps XYZ coordinates into two distinct values (\textit{i.e.}, the block index and the offset index), which represent the discretized 3D space separated by the global block and the local offset.
\revise{Furthermore, the vertices are integrated within each block, assigning the block index once per group in the sequence. This compresses the final sequence length to roughly match the number of vertices. Where the number of faces ($F$) is about twice the number of vertices ($V$), and each face in the vanilla mesh representation ($S$) requires nine tokens, we obtain $ S=9F\approx18V$. This reduction in sequence length significantly improves inference speed and stabilizes generation by reducing the accumulation of prediction error.}

Even though block-wise indexing efficiently compresses the sequential representation, its reliance on 3D space discretizations leads to the loss of fine geometric details. As the complexity of 3D objects increases and the number of vertices grows, higher spatial resolution becomes necessary, which further amplifies this degradation. 
This issue primarily stems from two factors: the shifting of vertex positions to quantized points and the reduction of points through the merging of neighbors. As shown in Fig.~\ref{fig:discretize}, discretizing an object with 7-bit resolution noticeably reduces geometric detail, while adjusting only the vertex positions to continuous values restores most of the original geometry. Motivated by this observation, we design the fidelity enhancer, a small transformer structure, that receives 7-bit discretized vertex positions and shape information to convert each vertex into a continuous coordinate by predicting the residual, as illustrated in Fig.~\ref{fig:pipeline}(c).

\subsection{Stage \textrm{II}: Face Generation}
\label{sec:stage2}
In this stage, we embed the vertices by using the bidirectional transformer to capture inter-vertex relationships and predict the edges to finalize the mesh structure. 
Since SpaceMesh~\cite{spacemesh2024} demonstrates the effectiveness of the spacetime distance function~\cite{law2022spacetime} for modeling relational structures, we adopt this function to compute interactions between vertices, defined as follows:
\begin{align}
\text{dst}(u, v) 
&= \text{dst}([u_1, u_2], [v_1, v_2]) \notag\\
&= \| u_1 - v_1 \|_2^2 - \| u_2 - v_2 \|_2^2
\end{align}
where dst$(\cdot)$ denotes the spacetime distance function, and $u$ and $v$ are input vectors, corresponding to embedded vertex features in our setting. Each vector is split in half, and the Euclidean distance is computed separately for the two parts. The final value is computed by subtracting the second distance from the first, enabling the representation of both positive and negative values. 
Instead of using the function as the final activation, as done in SpaceMesh, we employ a multi-head approach in which the feature vector is split and the function is applied independently across the heads, with each result forming the distinct dimension of the edge feature. 
This edge feature is then passed through a prediction network, which outputs the edge logits for determining vertex connectivity(see Fig.~\ref{fig:pipeline}). 
This strategy significantly improves the network's capacity to represent diverse and complex connectivity. Lastly, we threshold the logits \revise{at zero} to determine edge connectivity, then construct mesh faces by identifying triplets of vertices that are mutually connected.

\revise{In the training phase, we adopt the asymmetric loss~\cite{ridnik2021asymmetric} to guide edge predictions, emphasizing the scarce positives whose proportion declines as the vertex count grows. Additionally, we prioritize reducing false-positive edges that create holes, rather than pursuing class balance, to preserve geometric fidelity. The loss is defined as follows:
\begin{align}
\text{AsyLoss}(p) =
\begin{cases}
L_{+} = (1 - p)^{\gamma_+} \log(p), \\
L_{-} = p^{\gamma_-} \log(1 - p),
\end{cases}
\end{align}
where $L_{+}$ and $L_{-}$ represent the loss values for positive and negative samples, respectively, and $\gamma_+$ and $\gamma_-$ are scaling factors controlling how steeply the loss decreases as predictions become more confident, set to $0$ and $3$ in our experiments.
It is noteworthy that the face generation process is executed in a single feedforward, which can achieve fast inference while producing sophisticated results.}

\begin{figure}
    \vspace{3mm}
    \centering
    \centerline{\includegraphics[width=0.9\linewidth]{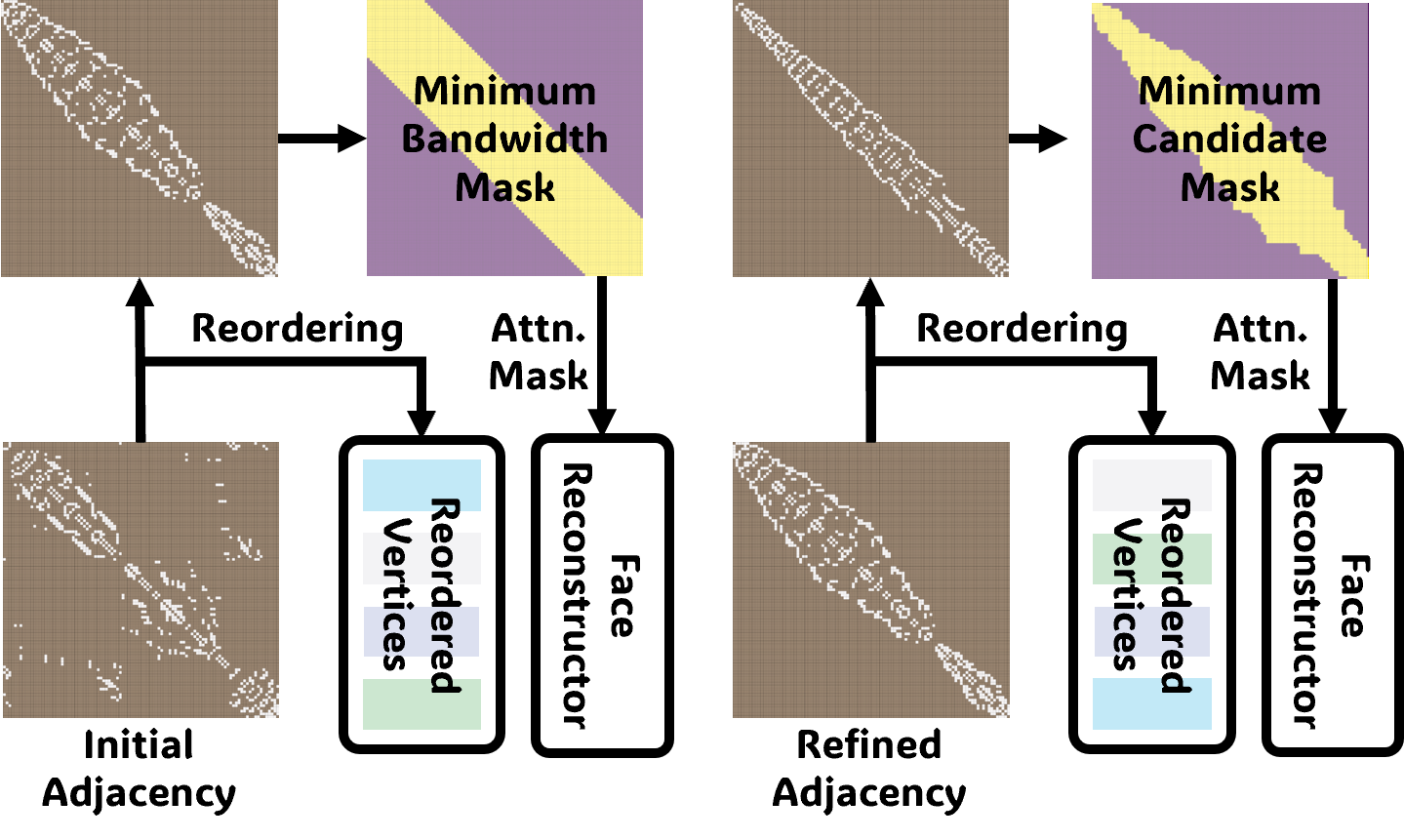}}
    \vspace{-2mm}
    \caption{\label{fig:candidate_reduction}
    The detailed structure of the prediction filtering. We use the initial adjacency matrix from the first face generation to perform BFS reordering. Based on this reordering, we apply the minimum bandwidth mask and the minimum candidate mask as attention masks.}
    \vspace{-3mm}
\end{figure}

\subsection{Post-processing: Prediction Filtering}
\label{sec:candidate_reduction}
\revise{Our face generation process occasionally predicts incorrect links, most of which do not significantly affect the mesh structure due to overlaps, but they increase computational cost. To address this, we introduce prediction filtering, a post-processing strategy applied after the initial adjacency matrix prediction.
The bandwidth of an adjacency matrix is defined as the maximum $|i-j|$ for which $A[i][j] \neq 0$, representing the farthest distance from the diagonal where a connection occurs. We first reorder nodes via breadth-first search to reduce bandwidth and construct a minimal bandwidth mask that retains edges with $|i-j| \le B$. Using this mask, we re-predict edges over two iterations, narrowing the candidate set at each step. Subsequently, we apply a minimum candidate mask, which assigns each node $i$ a maximum valid connection distance $r_i$ (retaining neighbors $j$ with $|i-j| \le r_i$), and repeat the two-iteration refinement. The entire process is illustrated in Fig~\ref{fig:candidate_reduction}. This two-stage, iterative filtering reduces unnecessary faces while preserving geometric fidelity.}

\begin{figure*}
    \centering
    \centerline{\includegraphics[width=1.0\textwidth]{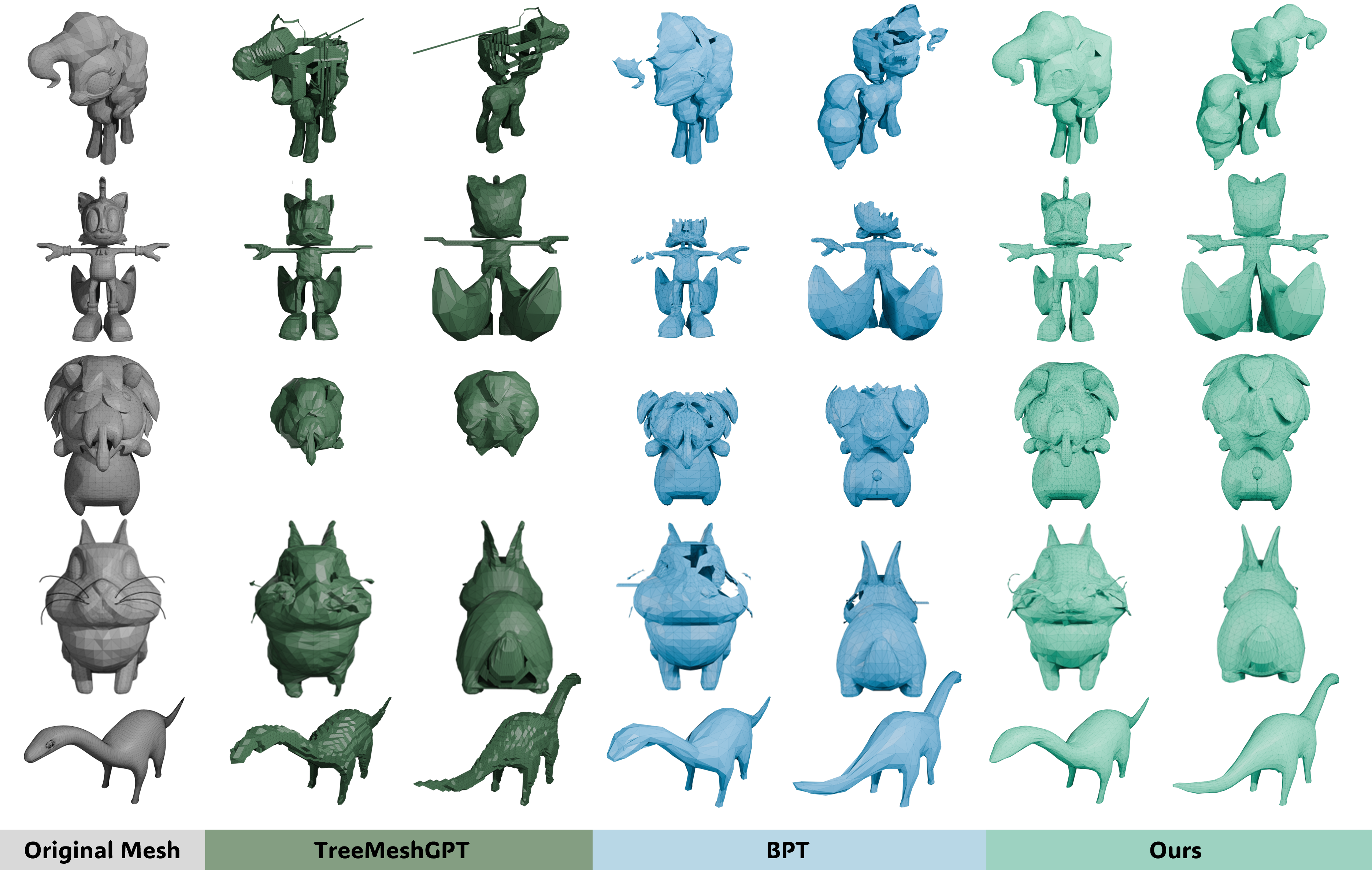}}
    \caption{\label{fig:comparision} Qualitative comparison of shape-conditioned mesh generation on the Toys4K dataset~\cite{toys4K}. All meshes were generated from the same input point clouds that were sampled from the original meshes.}
    \vspace{-3mm}
\end{figure*}

\section{Experiments}
\subsection{Implementation details}
\label{sec:Experiment}

\paragraph{Model Variants.}
We introduce two variants of our method, i.e., \model-V4K and \model-V1K. The two variants share an identical model structure, but differ in the filtering applied to the training data: \model-V4K is trained on meshes with up to 4,000 vertices, whereas \model-V1K is trained on a subset filtered to include only meshes with no more than 1,000 vertices.

\paragraph{Datasets.}
To train the full network, we combine ShapeNet~\cite{chang2015shapenet}, Objaverse~\cite{objaverse}, and a portion of Objaverse-XL~\cite{objaverseXL}, selecting samples with fewer than 4,000 vertices. We further filter out undesirable meshes with our algorithm and finally obtain a dataset of 100K high-quality meshes. 
For inference, we use the Toys4K dataset~\cite{toys4K}, which contains 4,000 meshes with a diverse range of complexity.

\paragraph{Evaluation Metrics.}
To quantitatively evaluate the quality of mesh generation, we adopt Chamfer Distance (\textbf{CD}) and Hausdorff Distance (\textbf{HD}) as evaluation metrics. Both the generated and ground-truth meshes are normalized in scale, and 5,000 points are uniformly sampled from each mesh surface.
CD measures the average closest-point distance between two point sets, capturing the overall structural similarity. HD captures the largest deviation by measuring the point that is farthest from any point in the other set, making it sensitive to local errors such as holes or incorrectly reconstructed regions.
Additionally, we calculate the average inference time (\textbf{Inf. Time}) and number of vertices (\textbf{\#V}) per mesh to demonstrate the efficiency of our framework. 
In the experiment for ablation studies, we also use the number of faces (\textbf{\#F}), F1-score, and recall.

\subsection{Performance Evaluation}
To demonstrate the effectiveness of our method, we conduct both quantitative and qualitative evaluations by comparing it with state-of-the-art approaches for shape-conditioned artistic mesh generation (\textit{i.e.}, MeshAnything~\cite{chen2024meshanything, chen2024meshanythingv2}, TreeMeshGPT~\cite{lionar2025treemeshgpt} and BPT~\cite{weng2024scaling}). 
For quantitative evaluation, we use the Toys4K dataset~\cite{toys4K}, which none of the compared models have used for training. In qualitative evaluation, we additionally utilize the samples from the ObjaverseXL dataset~\cite{objaverseXL}. The results for comparisons are conducted without using post-processing. 

\paragraph{Quantitative.}
\setlength{\tabcolsep}{2.5pt}
\begin{table}[]
\caption{Quantitative comparison on the Toys4K dataset. 
The best and second best results in each category are \textbf{bold} and \underline{underlined} respectively.
}
\centering
\small
\begin{tabular}{l|ccccc}
\hline
Method & CD $(\%)$ $\downarrow$ & HD($\%$) $\downarrow$ & Inf. Time(s) $\downarrow$  & \#V \\ \hline
MeshAnything   & 12.02  & 26.87  & 26.06  & 218.6   \\
MeshAnythingV2 & 10.23 & 24.98 & 31.94 & 533.3   \\
TreeMeshGPT    & 5.46  & 13.96 & 27.32 & 706.3   \\
BPT            & 5.71  & 12.02 & 49.23 & 525.5   \\ \midrule
\model-V1K     & \underline{4.09}  & \underline{10.32} & \textbf{3.41}  & 467.2    \\ 
\model-V4K     & \textbf{4.05} & \textbf{10.22} & \underline{6.60}  & 1040.6   \\ \hline
\end{tabular}
\vspace{-3mm}
\label{table:comparision}
\end{table}
For evaluation, we used the Toys4K dataset by selecting meshes with fewer than 5,000 vertices, resulting in a total of 2,063 samples. 
The comparison results are presented in Table~\ref{table:comparision}. As shown in the results, the \model-V4K achieves the best performance, with scores of 4.05 and 10.22 on Chamfer distance and Hausdorff distance, respectively. In terms of generation time per vertex, \model-V4K is eight times faster compared with BPT while having better geometry representation. Moreover, \model-V1K demonstrates the fastest inference time, with an average of approximately 3.41 seconds per mesh, which is nearly twice as fast as \model-V4K, owing to its use of fewer vertices.
It also shows that the stability of mesh generation by using such long sequences often fails due to the accumulation of prediction errors during generation, which further deteriorates overall performance. It is noteworthy that our proposed approach deals with shorter sequences that mitigate not only the inference speed but also the overall performance.

\begin{figure*}
    \centering
    \centerline{\includegraphics[width=1.0\textwidth]{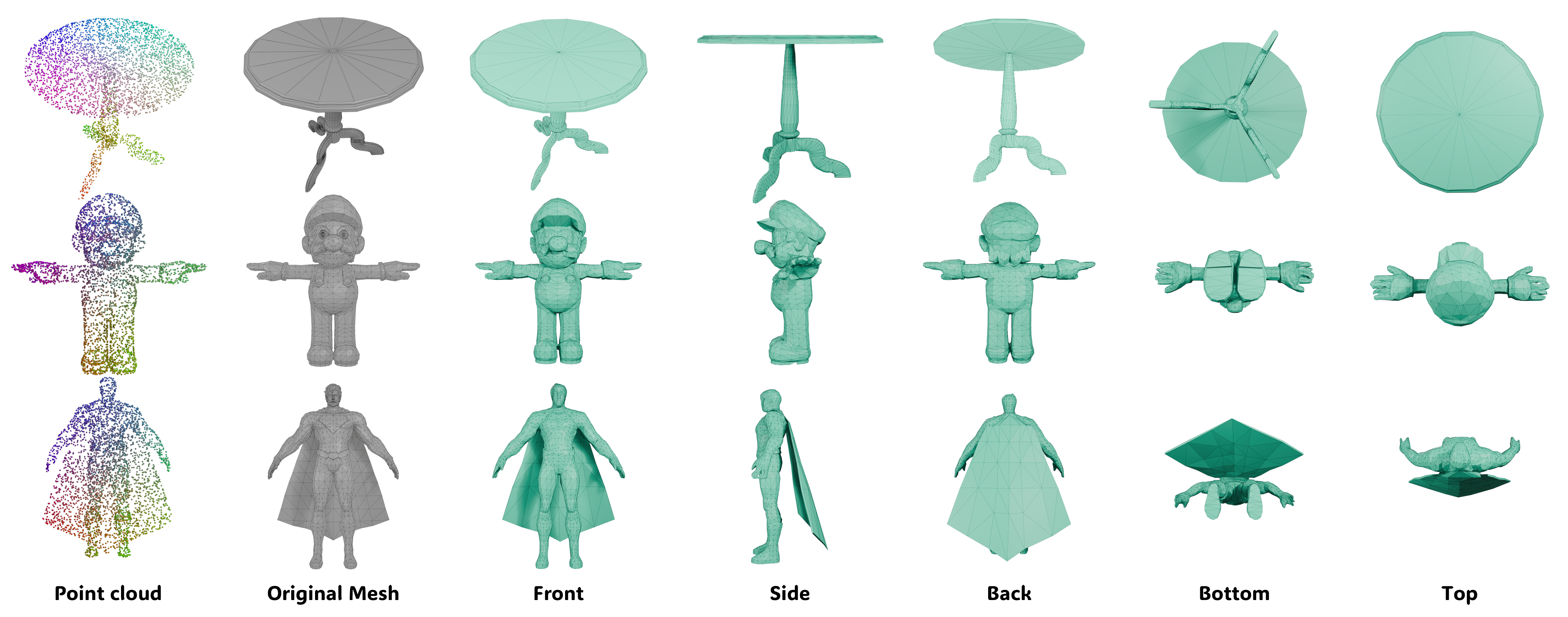}}
    \vspace{-4mm}
    \caption{\label{fig:objaverse} Diverse view of mesh results from the proposed method in the Objaverse dataset~\cite{objaverse}.}
    \vspace{-3mm}
\end{figure*}

\paragraph{Qualitative.}
We conducted a qualitative evaluation using high-resolution meshes from the Toys4K dataset, as shown in Fig~\ref{fig:comparision}. As illustrated, the proposed method consistently produces more accurate and refined meshes compared to other approaches. Notably, other methods often fail to finalize the mesh structure, resulting in incomplete or distorted shapes. In contrast, our method generates complete and coherent geometry, effectively preserving fine-grained features throughout the generation process. \revise{This is due not only to token-length constraints but also to the growing instability at longer sequences, as shown in the first example.}
There are also cases where the overall shape is preserved, but detailed features are not well represented. For instance, in the fourth row of the comparison with TreeMeshGPT, the rabbit exhibits simplified facial structures, particularly poorly defined eyes and unnaturally thin arms, leading to lower visual quality.
We also demonstrate the effectiveness of our fidelity enhancer through the example in the last row by showing smoother surfaces. 
Additionally, we present multi-angle views of our generated meshes across samples of varying complexity from the Objaverse dataset in Fig.~\ref{fig:objaverse}, demonstrating that our results are consistently well-structured.

\subsection{Ablation Studies}
In this subsection, we demonstrate the validity structure and loss function for face generation by the comparative experimental results according to the change of components. For the experiment, we train the model on the ShapeNet dataset and evaluate it on the Toys4K dataset by using meshes with fewer than 500 vertices in both.
Moreover, we demonstrate the effectiveness of the fidelity enhancer, which locates vertex positions in a natural arrangement.

\paragraph{Edge Prediction.}

To demonstrate the effectiveness of the structure for the final step in the face generation, i.e., predicting edge connectivity using embedded vertex features, we compare our method against four different designs: using spacetime distance function as final activation as used in SpaceMesh~\cite{spacemesh2024}; computing the function multiple times across split feature groups and averaging the results as logits; replacing the spacetime distance function to cosine similarity; and following the multi-head manner which use the MLP block to predict logits. The following results are shown in Table~\ref{table:edge_prediction}.
Based on the comparison between the first and second settings, the multi-head configuration provides a significant improvement over the case that uses a single vector. This indicates that a single vector is insufficient to capture all complex relationships between vertices. Furthermore, the design using the spacetime distance function outperforms the one using cosine similarity, reaffirming the effectiveness of the spacetime distance function in graph representation, as demonstrated in previous work.~\cite{law2022spacetime, spacemesh2024}.

\setlength{\tabcolsep}{7pt}
\begin{table}[]
\caption{Performance analysis of the face generation according to the change of the edge prediction structure based on the Toys4K dataset. }

{
\small
\begin{tabular}{lcc|cc}
\hline
Function & Multi-head & MLP  & CD $(\%)\downarrow$ & HD$(\%)\downarrow$ \\ \hline
Spacetime & \xmark & \xmark & 7.27  & 25.32  \\
Spacetime & \cmark & \xmark & 5.72 & 21.41  \\
Cosine    & \cmark & \cmark & 5.78   & 22.05  \\ 
Spacetime & \cmark & \cmark & \textbf{5.06}  & \textbf{18.55}  \\ 
\hline
\end{tabular}
}
\label{table:edge_prediction}
\end{table}
\paragraph{Loss Function.}
\setlength{\tabcolsep}{6pt}
\begin{table}[]
\centering
\caption{Performance analysis of the face generation according to the
change of the loss function based on the Toys4K dataset.}
{
\footnotesize
\begin{tabular}{c|ccccc}
\hline
Loss & CD$(\%)\downarrow$ & HD$(\%)\downarrow$& F1$(\%)\uparrow$ & Recall$(\%)\uparrow$ \\ \hline
BCE                  & 11.62   & 36.27   & 69.45  & 64.07 \\
Dice                 & 10.88   & 35.68   & 67.60  & 71.20 \\ 
Asymmetric           &  \textbf{5.06}   & \textbf{18.55}   & \textbf{70.58}  & \textbf{81.32} \\ \hline
\end{tabular}
}
\vspace{-3mm}
\label{table:loss_function}
\end{table}

We conducted a comparative analysis of loss functions to evaluate their importance in training a face generation model. We compared asymmetric loss~\cite{ridnik2021asymmetric} with binary cross-entropy, dice loss~\cite{milletari2016diceloss}, which are commonly used in binary classification tasks. In addition to standard geometry-based metrics, we also used F1 score and recall as evaluation metrics to better understand the relationship between the accuracy of matrices and the actual quality of the mesh.
The experimental results show that the asymmetric loss consistently outperformed the other loss functions across all metrics. Since the asymmetric loss forces the model to focus more heavily on learning from positive samples, it leads to more accurate predictions, despite the low proportion of positive samples in the matrix. Dice loss, which also emphasizes learning from positive values, showed improved performance in Chamfer Distance (CD) and Hausdorff Distance (HD) compared to BCE loss. However, it does not enforce this emphasis as strongly as the asymmetric loss, resulting in comparatively lower overall performance. Notably, although the F1 score remains similar across all loss functions, the model trained with asymmetric loss achieves a significantly higher recall with improved shape metric scores. This suggests that positive predictions contribute more critically to the representation of mesh structures, as improved recall is strongly associated with better shape metric scores.

\begin{figure*}[h]
    \vspace{-3mm}
	\centering
    \begin{minipage}{0.3\textwidth}
        \centering
            \vspace{2mm}
            \includegraphics[width=1.0\textwidth]{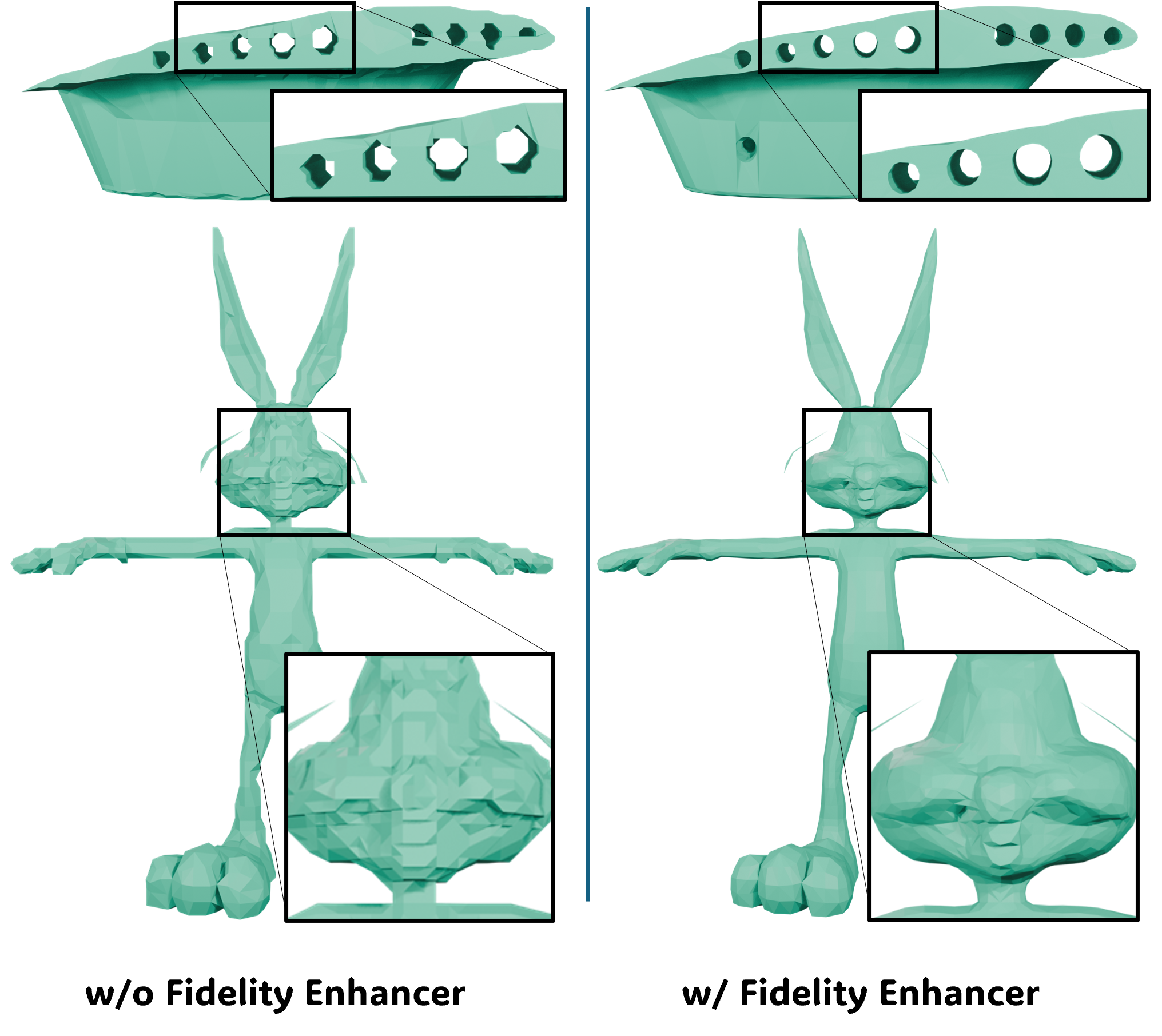}
            \caption{Comparison of the mesh results from \model-V4K according to the usage of the fidelity enhancer.}
            \label{fig:fidelity_enhancer}
    \end{minipage}
    \hfill
    \begin{minipage}{0.68\textwidth}
        \centering
    	\includegraphics[width=1.0\textwidth]{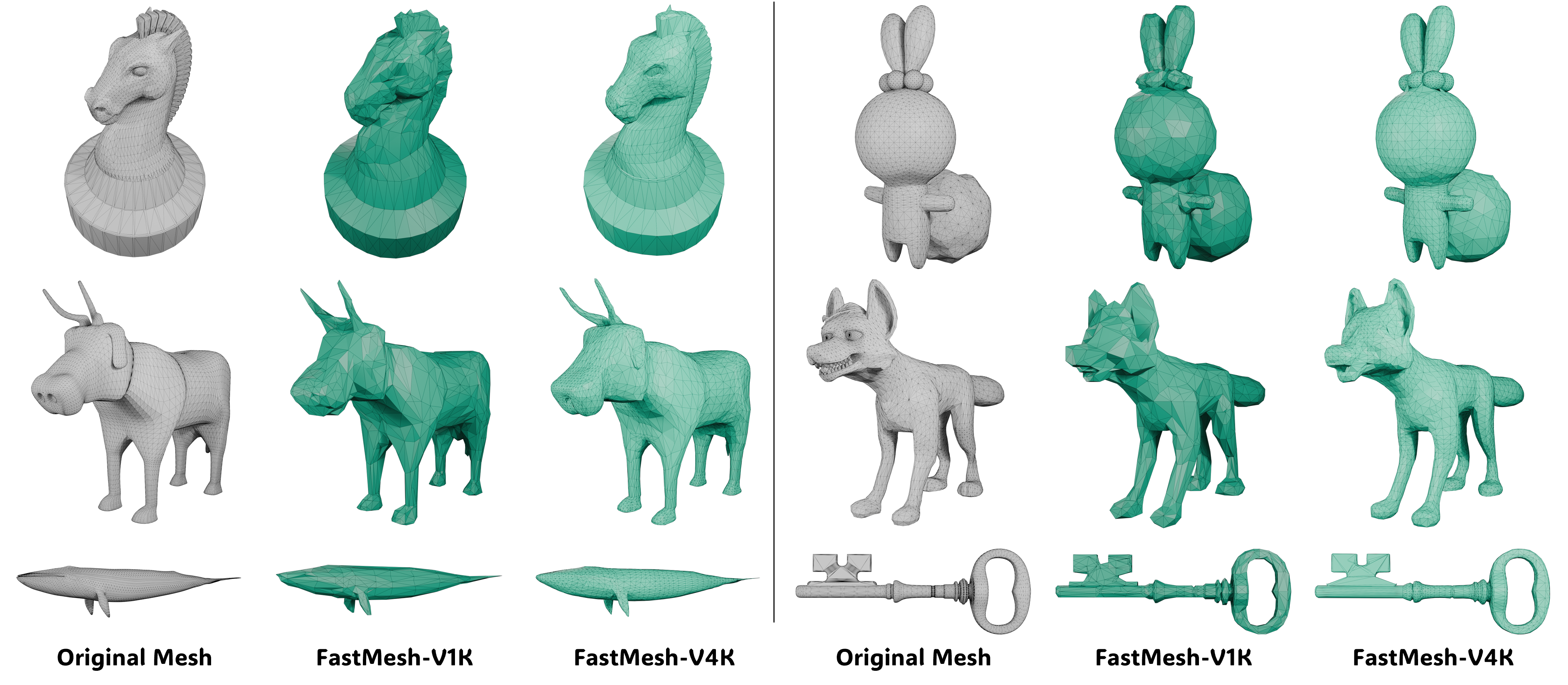}
        \vspace{-10mm}
            \caption{Comparison results of the two variants of the proposed method. The meshes are generated from input point clouds extracted from the original meshes for each model.}
            \label{fig:V1KV4K}
	\end{minipage}
    \vspace{-3mm}
\end{figure*}

\paragraph{Prediction Filtering.}
\setlength{\tabcolsep}{4pt}
\begin{table}[]
\caption{Performance analysis of face generation with and without prediction filtering (\textbf{PF}) on the Toys4K dataset. \#F and \#V denote the number of faces and vertices, respectively.}
\centering
\small
\begin{tabular}{c|ccccc}
\hline
PF  & CD$(\%)\downarrow$ & HD$(\%)\downarrow$ &  Inf. Time$(s) \downarrow$  & \#F & \#V \\ \hline
\xmark  & 4.05   & {10.22}  & {6.77}   &  6799.2 & 1040.6 \\
\cmark  & {4.03}   & 11.02  & 12.59  &  2811.1 & 1040.1 \\ \hline
\end{tabular}
\vspace{-3mm}
\label{table:prediction_filtering}
\end{table}

We conducted an experiment to evaluate how effectively unnecessary faces can be removed through prediction filtering. We compare with generation metrics along with the average number of faces and vertices in each result, as shown in Table~\ref{table:prediction_filtering}. As observed, the results without post-processing show that the number of faces is excessively higher than the number of vertices. Most of these extra faces are overlapping surfaces lying on the same plane. Although they do not affect the mesh geometry, they result in unnecessary computational overhead in downstream applications.
Although post-processing leads to a slight decrease in Hausdorff Distance due to removing and making holes in the failure cases, it improves Chamfer Distance while substantially reducing the number of faces. This indicates that the method effectively removes unnecessary faces without compromising geometric quality.

\paragraph{Fidelity Enhancer}
We evaluated meshes generated with and without the fidelity enhancer under identical conditions and found clear improvements in facial details and smoother surfaces (Fig.~\ref{fig:fidelity_enhancer}), demonstrating that our simple refinement network can significantly improve mesh quality by adjusting vertex arrangements.

\subsection{Discussion }
\label{sec:discussion}
\paragraph{Variants}
 We introduced two variants that exhibit different generation behaviors by training with different data filtering. \model-V1K is aimed at accelerating the process, while \model-V4K prioritizes \revise{mesh quality.} As shown in Table~\ref{table:comparision}, although \model-V1K handles significantly fewer vertices than \model-V4K, the performance in terms of Chamfer Distance (CD) and Hausdorff Distance (HD) remains comparable. This is because these geometry-based metrics are more sensitive to overall structural accuracy than to fine-grained details, and \model-V1K is still able to capture the mesh structure effectively with fewer vertices. As illustrated in Fig.~\ref{fig:V1KV4K}, even when given a complex structure, \model-V1K successfully reconstructs the overall geometry without structural collapse. On the other hand, \model-V4K processes a larger number of tokens, trading off some generation speed for higher detail. This allows it to generate meshes with greater vertex density, resulting in smoother and more elaborate surfaces.

\vspace{-1mm}
\revise{\paragraph{Limitations and Future Work}
While our method shows strong performance, several limitations remain. In vertex generation, the model can occasionally produce overly fine-grained sequences that exceed the maximum vertex limit. In face generation, it may remove valid faces, retain invalid ones, or produce overlapping faces, meaning manifoldness is not guaranteed. Details and examples are provided in the supplementary material. Future work will explore relative positional encoding~\cite{su2023roformerenhancedtransformerrotary} to allow longer sequence generation, refine the face generation architecture, and introduce constraints or additional pipelines to ensure manifoldness.}

\section{Conclusion}
In this paper, we propose a simple yet powerful framework for artistic mesh generation. 
We address the challenge of representing the mesh in a single sequence by treating the vertices and surfaces of the mesh separately. We enable the autoregressive model to process only a few tokens for the process, which results in faster inference speed and more stable results. To mitigate the discretization issue introduced by the vertex quantization process, we apply the fidelity enhancer, which refines vertex positions. Furthermore, we reduce the number of unnecessary faces through the prediction reduction process, making the results from our method to be in well-designed. 
Our experimental results demonstrate that \model delivers significantly more robustness across diverse shapes compared to prior methods, while generating highly detailed meshes up to 8$\times$ faster. 
\revise{We have shown that the decoupled approach offers an efficient solution for mesh generation, and we believe it will inspire further research in the field.}
\section*{Acknowledgements}
This research is supported by the National Research Foundation, Singapore, under its NRF Fellowship Award NRF-NRFF16-2024-0003. This research is also supported by NTU SUG-NAP, as well as cash and in-kind funding from NTU S-Lab and industry partner(s).

{
    \small
    \bibliographystyle{ieeenat_fullname}
    \bibliography{main}
}



\end{document}